# A Survey of Research on Control of Teams of Small Robots in Military Operations

Stuart Young, Army Research Laboratory
Alexander Kott, Army Research Laboratory

## Point of Contact:

Alexander Kott
Army Research Laboratory

E-mail Address: alexander.kott1.civ@mail.mil

## Abstract

While a number of excellent review articles on military robots have appeared in existing literature, this paper focuses on a distinct sub-space of related problems: small military robots organized into moderately sized squads, operating in a ground combat environment.

Specifically, we consider the following:
- Command of practical small robots, comparable to current generation, small unmanned ground vehicles (e.g., PackBots) with limited computing and sensor payload, as opposed to larger vehicle-sized robots or micro-scale robots;
- Utilization of moderately sized practical forces of 3–10 robots applicable to currently envisioned military ground operations;
- Complex three-dimensional physical environments, such as urban areas or mountainous terrains and the inherent difficulties they impose, including limited and variable fields of observation, difficult navigation, and intermittent communication;
- Adversarial environments where the active, intelligent enemy is the key consideration in determining the behavior of the robotic force; and
- Purposeful, partly autonomous, coordinated behaviors that are necessary for such a robotic force to survive and complete missions; these are far more complex than, for example, formation control or field coverage behavior.

## Introduction

In this paper, we explore a specific sub-space of robotic command and control: small military robots organized into moderately sized squads operating in ground combat environments.

For the sake of illustration, we introduce an example of a specific mission that we will be using throughout the paper. Consider an infantry platoon that has the mission to clear and secure several three-story buildings. The platoon includes a small force of three partly autonomous small robots (sized like modern Small Unmanned Ground Vehicles [SUGVs] (*1*)) and five unattended ground sensors (UGSs).

Normally, once it has cleared a building, the platoon moves to the next building, leaving behind a squad of Soldiers to ensure the cleared building remains secure. In this paper, we call this task the "secure-building mission." Manning the secure-building mission greatly diminishes the remaining strength of the platoon. As an alternative, the platoon could leave behind one or two UGSs (typically equipped with acoustic sensors only) to continue monitoring the building. However, an intelligent adversary could easily spoof or defeat such simple sensors.

Imagine, however, instead of leaving behind a squad of human Soldiers, the platoon leaves behind 3–5 small robots (similar to the Future Combat System [FCS] SUGV-class robot) and 5–10 small stationary sensors, such as modern UGSs, e.g., AN/GSR-10 (V) 1 (*2*) or McQ iScout (*3*). Assuming this combined force is capable of performing the collaborative, autonomous secure-building mission, the platoon could then move on to its next mission objective with its full complement of personnel while the robotic force maintains the security of the cleared building.

We propose this example mission as a suitable challenge problem to the small-robot community. This challenge problem is particularly useful for several reasons. First, this type of mission is highly practical and ready applies to real-world operations. Second, this mission combines numerous challenging technical aspects in one well-defined task. Third, it is relatively easy to experiment with this task. Finally, we view this challenge as one that should be solvable in near-term; our team is currently pursuing solutions to this problem.

This paper's sections correspond to key applicable areas of robot squad capabilities. The first section explores the ways in which the squad may acquire awareness of its surroundings, including the physical shape and features of building and relevant friendly and hostile forces, as well as how individual members of the squad integrate that awareness and how to combine that awareness with the commander's guidance. We then review how the robotic squad might communicate among its robotic assets and with humans, including the controller of the squad. Then we cover how the robotic squad would make decisions while planning and executing its mission, and how it might produce desired effects on the enemy and non-combatants.  We point out the challenges associated with command and control (C2) of this squad, including the fact that the C2 of such forces differs significantly from the C2 of human forces and larger robots. We also look at ways to integrate a robot squad into the C2l structure (e.g., Fig. 1), either traditionally or atypically.

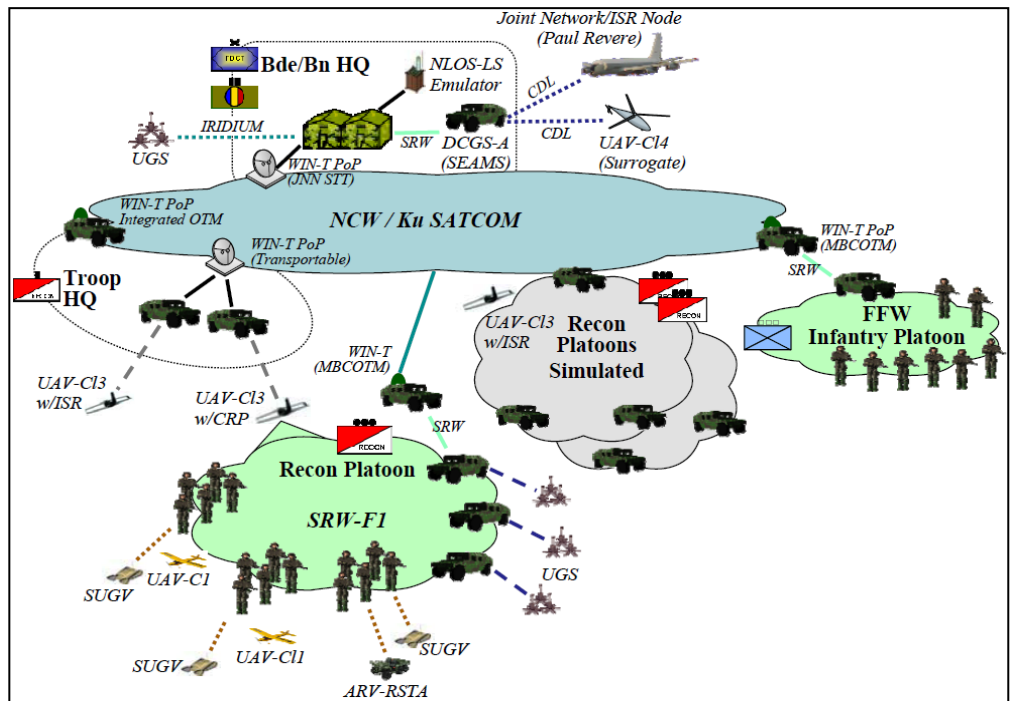

Figure 1. This experimental Command and Control (C2) architecture incorporates small teams of heterogeneous robotic assets. The concept has undergone extensive experimental evaluation (*4, 5*).

## Section1. Perception and Situational Awareness of the Small-robot Squad

In executing the secure-building mission we introduced previously, the squad of small robots faces a number of challenges. The robots need an understanding, a model of their physical environment.  In the secure-building mission, the model could be the layout of the building including its 3-D structure, obstacles, windows, doors, and walls.  In addition, the model would need to incorporate the location of each robot within the environment. The robots would also need to know where potential Red personnel are located in the building and how to recognize them as potentially hostile. They would also need to detect and recognize non-combatants and friendly forces.

To achieve this level of awareness, a team of small robots and sensors faces a number of challenges: the relatively inexpensive and lightweight sensors and processors available to small robots, the field-of-view limitations typical to a small robot in a three-dimensional (3-D) environment, and the necessity of a collaborative formation of situational awareness.

## Section 1.1 Awareness of Physical Environment

The Sick laser detection and ranging (LADAR) sensor has become the de facto standard for providing ranging information to ground robotic platforms in academia and larger military systems. In particular, the experiences of the DARPA Grand Challenge and Urban Challenge (*6*) have positioned the Sick LADAR as a nearly universal choice for the best-performing autonomous cars.

There have been attempts to use the Sick LADAR for small robots. For example, the Navigator payload (*7*) for a PackBot Scout includes a Sick-LD OEM LADAR, an Ublox GPS, a 3DM-GX1 IMU, a KVH gyro, and a serial radio. The Navigator derives from the Wayfarer project at iRobot (*8*). Similarly, the FCS autonomous mobility sensor (*9*), part of the Autonomous Navigation System (ANS) program (*10, 11*), has a 3-D LADAR that is very capable, but is still approximately the size of the Sick. These sensors, however, are far too large for a SUGV.

One alternative to a Sick is the Swiss Ranger SR-2 (*12*). This flash LADAR differs from traditional LADARs in that it uses pulses of infrared (IR) light to determine ranging information, rather than a scanning laser. It has a relatively small form factor (compared to traditional LADARs) and provides 3-D ranging information. However, the sensor does not operate reliably outdoors. Another alternative is the Hokuyo URG-LX (*13*), a traditional two-dimensional (2-D) scanning LADAR with a form factor of roughly one quarter the size of the Swiss Ranger. Although the Hokuyo's performance diminishes in bright, sunny conditions, it performs well under indoor conditions and is practical under relevant missions, such as building surveillance.

LADAR-based near-real-time mapping of physical environments also offers an approach to collaborative mobility. In this approach, an airborne robot generates a high-resolution 3-D terrain model that its ground-based counterparts can then use for effective navigation (*14–16*). This approach greatly simplifies the challenges of obstacle detection and avoidance.

Given the limitations of LADARs suitable for small robots, the stereo-imaging approach to generating a 3-D terrain model may be an attractive option, even though it is becoming rather unpopular for larger robots (*17*). For example, the Seer payload (*7, 8*) developed by iRobot incorporates a Tyzx stereo camera head that replaces the PackBot Scout head. The 3-D stereo data provided by the sensor enables 3-D obstacle avoidance, visual odometry, and 3-D modeling. Typically, in a stereo-imaging approach, the data from the stereo camera is transformed into a 3-D point cloud, which is then used to compute the probability of finding obstacle in a given 3-D subspace (*18*). Besides stereovision, monocular and omnidirectional sensors have also been used to determine robot location (*19*).

Integration of LADAR- and stereo-based perception could be a particularly promising approach. In a typical example of partial-autonomy (*20*), a Packbot is equipped with stereo cameras and a single axis scanning LADAR, which together allow a variety of behaviors ranging from teleoperation to waypoint following with obstacle avoidance.

A significant practical questions remains: Can a minimal sensor suite be composed that provides adequate robot performance to accomplish the mission given variations in the operational complexity and environment (*21*)?

## Section 1.2 Awareness of Friendly and Enemy Forces

Perception of friendly forces is critical to avoid fratricide. The robot squad could accomplish this critical task by using approaches similar to those currently applied to Blue force tracking (*22*). Systems such as Land Warrior (*23*) currently provide Soldiers with Blue force tracking information. Adding this technology to the robots would make them part of the "Blue force" picture, thus providing situational awareness of friendly Soldiers and friendly robots.

Other techniques, such as Simultaneous Localization and Mapping (SLAM) (*24*), provide robots alternative means for determining their locations in GPS-denied environments. Knowing the location of other friendly robots is necessary to enable collaborative behaviors among robots. Other technologies, such as ultra-wide band (UWB) (*25*), could also be incorporated into the system to provide the location of friendly forces and robots.

One approach to enemy detection and targeting uses acoustic and video shooter detection techniques, which have seen rapid advances (*26*, *27*). The detection of acoustic events, such as gunshots from a robotic platform, has been demonstrated for many years (*28–31*). In addition to acoustic techniques, a number of approaches exist for the detection of humans using computer vision (*32*, *33*). The primary remaining issue is achieving an acceptable probability of detection with acceptable false alarm rates.

## Section 1.3 Integration of Awareness

To integrate a broad range of sensors, controls, software modules, and external communication signals, a SUGV requires a flexible robot software architecture. The architecture should work across multiple heterogeneous platforms, independent of the robot geometry, and must accommodate diverse sensors and behaviors. Two examples of such an architecture are the INL Robotic Intelligence Kernel (RIK) (*34*) and the Autonomous Capabilities Suite (ACS), developed by SPAWAR Systems Center (SSC), San Diego (*35*). ACS and RIK each allow rapid integration of new payloads, perceptions, and behaviors by abstracting and publishing all data in a standard and easy-to-use manner. They both also allow the robot to recognize the sensors available at any given time and adjust its behaviors and perceptions accordingly.

Another flexible software architecture is the National Institute of Standards and Technology (NIST) Mobility Open Architecture Simulation and Tools (MOAST)—an open-source, modular implementation of the Four-Dimensional/Real-Time Control System (4D/RCS) architecture developed for the Urban Search and Rescue competition (*36*, *37*).

Data sharing between several robots is a challenging and important capability for a small-robot squad. In particular, each robot in the squad needs accurate information about its own location and the locations of other squad robots with respect to each other and within the building. Localization can be difficult in the environment of the secure-building mission, where GPS, INS, and landmark identification all face considerable obstacles. In such cases, robots need to collaborate to help each other to identify their positions more accurately (*38–40*).

Collaborative mapping can also assist in constructing an integrated awareness. While mapping and SLAM by a single robot are extensively researched topics, combining and deconflicting maps (especially 3-D maps and maps obtained from robots with unreliable localization) remains a difficult problem (*24*, *41–43*).

The topic of integrated awareness brings up a key issue related to C2: The controller of the robotic squad must be able to provide the squad with his own awareness of the situation. There are several issues associated with this problem. A human's perception and representation of the situation is likely to be more abstract (e.g., "you need to watch a shed in the back") than that of a robot. Abstract representation formulated by a human is very difficult to merge with robot conceptualization. For example, identifying the "shed" within the building map constructed by the robots can be problematic, especially if the robots have yet to construct the map when the controller issues guidance to the squad. Similar difficulties apply when communicating human understanding of the enemy and non-combatant situations to the robots.

## Section 2. Communications and Interactions with Other Robots and Humans

Small robots operating in urban and mountainous terrain face a number of special challenges, for example the highly unreliable communications typical in such environments. Difficulties resulting from this problem include the following:
- The need for concise and inexpensive communications and autonomous action if subsequent communications fail;
- The interdependence of movement planning and communications, i.e., ways to improve the robots' chances for communications by moving to appropriate points;
- The interdependence of survivability and communications to limit the chance of detection by the enemy; and
- The need to minimize the cognitive load of a human controller.

### *Section 2.1 Communicating with Robots*

Robots within a squad can communicate with each other (and with humans) in two paradigms. One method is the explicit exchange of messages through radio frequency (RF) links. The second method is stigmergic, i.e., a robot obtains information from another robot by merely observing the clues another robot left in the environment or observing the other robot's actions (*44*). Both paradigms are tested in the secure-building task. Robots patrolling different floors of a building, or different wings of a floor, may not be able to establish reliable, frequent RF contacts. Additionally, the robots may not often establish visual contact with each other. Thus, a combination of the two paradigms may be beneficial. For example, to minimize unsuccessful attempts at RF communications (wasting energy and providing clues to the enemy), a robot could initiate an RF link (explicit communication) only when it has visual contact with the other robot (stigmergic communication).

Communications between robots significantly improve collective performance but need not be complex (*45*, *46*). Communication languages (*47*, *48*) should be frugally adapted to the mission. For example, in the secure-building mission where robots may have infrequent opportunities for communications, the important information is knowing who does what and when. The language should allow description of the area the robot plans to patrol, the sequence or plan of the patrol, and the likely location and time window when contact will be initiated again. The collaborative decision-making paradigm applied within the robot squad (see section 3) also affects the nature of intra-squad exchanges.

A related, practical challenge is the need for application-agnostic, platform-agnostic infrastructure for multi-robot communications. One example of such an infrastructure is the Agile Computing Infrastructure (ACI) (*4, 5*). ACI is middleware that serves to isolate and protect applications from changes in the underlying communications infrastructure by providing a layer of abstraction that allows the applications to communicate seamlessly and transparently across a variety of devices. This infrastructure allows various robotic control requirements (teleoperation, telemetry data, and streaming video) to persist at a sufficient level of service in inherently unstable battlefield network environments and on any available network infrastructure or technology. ACI also allows the producers of information to advertise the type of service they provide, and subsequently allows consumers to locate a source for the type of service they require. Producers and consumers can then initiate and carry out their communications using the ACI. In the event of an extended network outage, automated service rediscovery can take place. Such an infrastructure is particularly relevant to operations in urban environments, where short-range communications may be temporarily interrupted causing assets to leave and rejoin the network frequently.

## Section 2.2 Interacting with Human Controllers

In addition to the communications challenges, interactions between robots and humans (particularly with the human controller of the squad) are complicated by fundamental differences in human and robot reasoning and representation. Simply put, humans do not think in the same terms as robots.

While there is extensive literature on human-robot interactions where one or several humans control a single robot, in this paper, we are focusing on how a single human issues high-level commands and control a squad of robots. To satisfy this purpose, two primary paradigms are emerging: sequencing and playbook (*49*). In the sequencing or switching paradigm (*50*), the human controller assigns individual tasks to each individual robot (e.g., "Patrol along these waypoints") in sequence, and then adjusts their coordination and timing as needed. In a playbook paradigm (*51*, *52*), the human selects a "play" (a plan of coordinated actions) from a pre-encoded collection of actions, adjusts appropriate parameters of the play, and commands the robot squad to execute the play.

Combining these two paradigms is also possible, e.g., begin with a play from a playbook, and then occasionally re-task an individual robot if circumstances require a deviation from the selected play. Alternatively, the controller can use a delegating approach (*53*), in which the human reviews the tasks of the play and then decides which one to control in a sequential manner and which ones to delegate to the robots to execute.

Another approach to controlling a team of robots (including a heterogeneous team) is to apply policy-based control, such as KAoS (*54*). Languages such as Web Ontology Language (OWL) provide a descriptive logic foundation that can be used to express semantically rich relationships between entities and actions, thus creating complex context-sensitive policies. KAoS includes a tool to create policies using OWL and an infrastructure to enforce these policies on robots. A policy-based approach may also be a more natural way for a human controller to manage multiple robots (*55*).

In the case of a secure-building mission, the robot squad controller is likely to be a dismounted Soldier. Most likely, he has to continue to fight as a member of the platoon and has many critical duties besides supervising the squad of leave-behind robots. Since the Soldier can only allocate limited attention to the robot-control duty, the design of human-robot interactions and the physical interface must take this limitation into account. For example, the Birddog human-robot-interaction console (IMU, safety and trigger switches, and communications suite mounted on an M4A1 assault rifle) (*56*) enables a robot to interpret the actions and intentions of humans based on their location and weapon status.

Another approach is to apply a multimodal interface on the dismounted Soldier that could consist of a weapon controller, a vibro-tactile belt, and an instrumented glove (*57*). Such an interface could lessen the cognitive load of the human controller and minimize physical actions beyond normal movement.

In general, increased autonomy of a robot significantly reduces the need for human interaction and cognitive load (*20*). However, the extent of the autonomy's benefit strongly depends on the user's training and experience (*34*). Using unmanned autonomous vehicles in both military applications introduces a new layer of cognitive complexity for human controllers (*58*). Designers of a robot squad must consider the complex relationships between situational awareness, controller utilization, and controller performance (*59*).

# Section 3. Individual and Collaborative Planning and Decision-making

There are three commonly accepted paradigms for describing the technique robots use to make intelligent decisions: hierarchical, reactive, and hybrid deliberative/reactive (*60*).

The hierarchical paradigm is historically the oldest. One example is the NIST Real-time Control Architecture (RCS) (*61*). The hierarchical paradigm is built on cycles of sensing the world, modeling the world, planning actions to achieve some goals in the world, and then acting to execute the plan. This process is often referred to as the SENSE-PLAN-ACT loop. In a multi-robot context, the hierarchical paradigm is often implemented in a centralized fashion, and often employs social analogies, such as roles and tasks allocation in an organization. The hierarchical paradigm has become less popular in recent years in favor of hybrid architecture.

The reactive paradigm links sensed patterns with acting behaviors, generally avoiding modeling and planning (*62*). It uses a SENSE-ACT organization of primitives. The subsumption architecture has been an influential example of the reactive paradigm (*63*, *64*). In a multi-robot context, the reactive paradigm is normally implemented in a decentralized fashion and the behaviors are often inspired by biological analogies.

Numerous multi-robot architectures have been proposed, most of which combine the deliberative and behavioral paradigms in various fashions, forming a hybrid (*65*). A few examples of hybrid architectures are ALLIANCE (*66*), Layered Architecture (*67*), Campout (*68*), and AuRA (*69*). The NIST RCS has evolved into a partly hybrid paradigm (4D-RCS) and is being used in the Army's FCS Autonomous Navigation System (ANS). Also, there is an open-source version called MOAST (*70*).

## Section 3.1 Applying a Deliberative Paradigm

In the secure-building mission, one may envision that the squad has a chief robot that acts as a centralized performer of the deliberative modeling, planning, and monitoring-replanning processes. The chief robot receives the mission from the human controller, formulates the plan of execution through tasks ("Task1: patrol second floor," "Task 2: monitor the east side of the front yard," etc.), allocates the tasks to the appropriate subordinates (robots and sensors), receives observations and reports from the subordinates, updates the model of the situation, monitors execution, and continually revises the plan, issuing the updated commands as necessary. This approach is similar to the C2 for the DARPA program HART (*71*), which deals with a team of primarily aerial unmanned assets.

From the C2 perspective, the deliberative paradigm offers several advantages to the human controller: the controller can easily understand this mode of operation, supply his own partial or complete plan, and review, approve, or modify the plan, etc.

However, the deliberative paradigm does include several challenges. For example, centralized planning and allocation of tasks is difficult in an environment with unreliable and infrequent communications. Further, each robot must react frequently and rapidly to unexpected events, such as stumbling onto a pile of rubbish in a room or an appearance of another robot or a human figure. Also, the deliberative paradigm places a heavy computing load on the chief robot and is subject to computational and communications lag.

Market-based techniques (*72*) combine the strengths of centralized and decentralized approaches, and can be implemented with a small-robot squad, especially when applied in a peer-to-peer fashion. Suppose Bot-1 is patrolling the first floor of the building while Bot-2 patrols the second floor. They meet in the stairwell and decide who will patrol the stairwell. Bot-1 says it would cost him 15.3, Bot-2 says it would cost him 12.7. The robots collectively agree that Bot-2 should patrol the stairwell. From the C2 perspective, however, market-based behavior is harder to understand and to control.

## Section 3.2 Applying a Reactive Paradigm

Alternatively, the robot squad can execute the secure-building mission using a reactive paradigm. For example, the robots, in ant-like fashion, can be given "virtual pheromones" (*73*) that would combine to guide each robot to (1) the least recently visited location, (2) locations where the robot could establish communications with another robot, and (3) locations of the most likely intrusion paths (see section 3.3). The resulting emergent behavior could provide effective results.

Unlike the deliberative paradigm, the reactive paradigm (typically associated with pattern-triggered behaviors, and often biologically inspired) does not require a centralized intelligent node, requires less computational resources (very important for small robots), allows robots to act rapidly in a changing situation or in response to sudden threats or opportunities, and can operate robustly in communications-starved environments.

The reactive paradigm also has several disadvantages. Reactive behaviors can be naïve at times and could be exploited by an intelligent adversary. It is also difficult to anticipate what behaviors will emerge, especially in a collective behavior of the squad in unusual or unanticipated circumstances. Swarming behavior (*74*) is a much-discussed area of research, but is not well suited for this task as the number of robots is quite small and each must focus on a well-circumscribed task. From the C2 perspective, reactive emergent behaviors are difficult to direct and control.

## Section 3.3 Applying a Hybrid Paradigm

A hybrid deliberative/reactive paradigm addresses many of the deficiencies with the two previously described paradigms. Adding a layer of supervisory control or a planning component to a reactive paradigm results in a less computationally expensive, reactive system that is not as naïve as a reactive system alone. Hybrid architectures are often broken up into basic modules (e.g., mission planner, mapping, etc.), of which some modules are distinctly deliberative and some are reactive (*60*).

Individual members of multi-agent teams can be programmed with behaviors following either the reactive or hybrid deliberative/reactive paradigms. Under the reactive paradigm, the multiple behaviors acting concurrently in a robot lead to an emergent behavior. In a multi-agent system, the concurrent but independent actions of each robot lead to an emergent social behavior. For homogeneous robots, a swarm approach may be implemented (*74*); however, for a heterogeneous team, a marsupial approach (*75*) may be more appropriate.

## Section 3.4 Reasoning in an Adversarial Environment

A small-robot squad must be mindful of the enemy while making decisions about securing the building. The importance of operating within the presence of an intelligent enemy has not been given enough attention in robotic literature. The robotic squad can fulfill its mission only by explicitly considering enemy actions and counteractions, using terrain to avoid detection and weapon fire from the enemy, and planning and executing actions in a way that maximizes chances for success in spite of intelligent efforts by the enemy.

Although the problem of multiple robots collaborating to patrol a designated area has been studied in various contexts using different approaches (*76*, *77*), most of this work does not explicitly take into account an intelligent adversary (an intruder who can observe the patrolling robots and devise methods of evasion). A few studies that do explicitly consider an intelligent adversary (*78*, *79, 106*) show that considering an adversary significantly complicates the problem.

The DARPA RAID program specifically focused on computational techniques of adversarial reasoning—estimating the current and upcoming actions of the opponent (*80*, *81*). Some of the algorithms developed in the program (*73*) are relatively inexpensive computationally and could be adapted to the limited CPUs available on small robots. In our secure-building problem case, the output of the algorithm would be the likely infiltration routes into and through the building. Each robot in the squad could use these estimated routes when forming patrolling decisions.

## Section 4. Producing Effects on the Environment

There are several ways in which a squad of robots can deliver effects or contribute to delivery of effects:
- Observing, detecting, and tracking targets and areas of interests;
- Threatening or delivering lethal and non-lethal fire;
- Designating targets; and
- Modifying enemy or non-combatant behavior by their presence at effective positions.

### Section 4.1 Observing, Detecting, and Tracking Targets and Areas of Interests

Mapping the building and identifying potential hazards are two tasks that a small robotic team may perform during the secure-building mission.  Booby traps and trip wires are known threats designed to deny access, slow troop movements, and inflict casualties.  Consequently, Soldiers take great care to avoid booby traps.  In addition, larger robots operating in the urban environment are much more expensive assets and thus may be designed with an element of self-preservation.  Small robots, on the other hand, are affordable assets that may be intentionally sacrificed to expedite mission success or may be directed to take greater risks to detect threats by getting closer to the threats, thereby making their sensors more effective (e.g., explosive sniffers) (*82–85*).

Another robot squad mission task could be long-endurance surveillance: robots could gather intelligence over a long period of time, reducing the burden on Soldiers.  When Soldiers perform such missions, their perception can wane when nothing happens in an area day after day, month after month. Some call this phenomenon the "dull" mission. Robots are ideally suited for this task, particularly in restricted areas (*86–89*).

One direction in persistent surveillance is applying computer-vision techniques to detect human intruders and peculiar activities (*90-92*).  Using advanced computer-vision algorithms on small robots may require improved robot architectures based on parallel processing to make these approaches computationally feasible.

To enable the observation, detection, and tracking of targets, robots may carry and intelligently deploy small sensors throughout the operational environment, in a marsupial fashion (*75*).  Adding deployable sensors to the collaborating small robots creates a reliable, fault tolerant, sensor network that can be remotely deployed, monitored, and tasked by human controllers and other assets.

### Section 4.2 Designating Targets and Applying Effects

Merely by their presence, robots may modify the behaviors of opponents and non-combatants, e.g., observing robots in an area may deter an enemy combatant from entering it. This function does not necessarily require an armed robot (*93*).

It also may be possible to use small robots as target designators, even in semi-autonomous mode. Given an accurate GPS location of a hostile fire source (*26*, *27*), a robot could employ GPS-based techniques for aiming a laser designator. For example, if a team of collaborating robots identifies the location of a sniper, the same robots could designate this point and enable a Soldier (after performing appropriate visual confirmation) to remotely deploy effects to the sniper's location.

Installing weapons on small robots (*94–96*) is a hotly debated and contested issue. Many are concerned with legal implications and the impact of collateral damage (*97*, *98*). Others argue that properly designed and commanded robots can be more ethical and more compliant with Laws of War and Rules of Engagement (ROEs) (*99*), and can significantly reduce collateral damage (*100*) as compared to human warriors.

## Section 5. Integrating a Robot Squad into a C2 Architecture

Fig.1 depicts a possible approach to integrating a squad of robots in to an Army brigade C2 structure. Note the close integration of several types of robotic assets: small ground mobile systems (SUGVs), small aerial platforms (UAVs), and stationary assets (UGSs). All these assets are organic to a platoon and distributed in small groups to squads of Soldiers. This particular architecture underwent extensive experimental evaluation (*4, 5*).

According to the experimental findings, such C2 structures provide Soldiers with an increased situational awareness. For example, information acquired by robots—such as imagery—becomes available to anyone who desires to obtain it via SPOT reports. Multiple robotic and human assets can collaborate across multiple networks at several echelons (*101*).

A less encouraging experimental finding in such C2 approaches is the need for what some call a Robotic NCO—a Soldier responsible for the robots. Experiments indicate that robot control (and the associated pre-mission planning) require significant level of training and expertise, and require a dedicated expert Soldier (*102*). The Robotic NCO tasks are cognitively demanding and require specialized training and development.

To reduce the need for such specialized command relations and expertise, and to increase C2 flexibility, researchers have explored potential architectures with significantly greater autonomy of robots. For example, Jacoby (*103*) argued that agile military C2 requires more distributed, autonomous approaches where robotic squads can execute partially autonomous teaming and swarming. Others looked at groups of robots with significant degree of self-management and explored formalism for the policy-based control of the robot squad (*104*). They propose to specify a mission in terms of roles, tasks, and the policies for managing tasks.

Yet others (*105*) have focused on many-to-many command relations between humans and robots. Instead of increased autonomy of robots, they aim to break the strict linking between owning an asset, commanding an asset, and making use of an asset. They are looking for ways to allow multiple networked authorities to take control and make use of the same robotic asset.

## Conclusions

Key technical solutions are emerging that allow robots to plan an effective tactical course of action; collaborate as situation demands; maneuver rapidly and effectively in a complex, dense, and dangerous environments; and know where the friendly forces are, find the enemy, and cause the desired effects. However, effective C2 of such robotic forces present significant challenges.

In particular, robot perception and situational awareness are fundamentally different from that of humans. A human controller or a robotic squad may find it difficult to adjust to such differences, and to appreciate that the robotic squad may perceive certain things much better than a human and others much worse. Communicating the commander's understanding of the situation to the robotic squad can be very problematic, and the extent of such a communication can be very restricted.

Further, the concepts of command via either a playbook, sequential/switching, or selective delegation are not particularly similar to command structure of human warriors. These methods of command call for a degree of precision and complexity that can be burdensome to the human commander. At least in the near-term, robotic squads are likely to be limited to a few suitable missions, such as the secure-building mission we used as an example in this paper. Understanding the possible variations and limits of such acceptable missions may be challenging for human commanders.

On the other hand, unique opportunities also emerge. If needed, ROEs can be rapidly changed and disseminated in a more agile fashion to robots than would be possible to human warriors in the field during an execution of an operation. Also, if required, re-tasking can be frequent and rapid, much more so than would be acceptable when commanding human warriors. Finally, detailed coordination between robots can be more precise and minute than would be appropriate to the cognitive abilities of human warriors.

Having received a human command, robots may proceed to make a lower-level execution decision in a manner that the human commander may find counterintuitive. The way robots reason, in many cases, is not analogous to the way a human reasons, even when adequate and appropriate for the robots' mission.

The tactics of a robotic squad are also likely to be different from those of human warriors. These non-humanoid tactics should take into account the unique strengths and weaknesses of robotic warriors. A robotic squad might use sacrificial tactics, accepting a higher degree of danger to the robot, in order to explore a dangerous target or draw fire of a concealed enemy. Robots can also execute a task in a highly uncomfortable environment with high endurance.

Command of robotic forces also raises complex legal and ethical issues that are yet to be fully understood and articulated. While some are concerned about using robots in any situation that may cause harm to humans, even indirectly (e.g., by surveillance of a target), others argue that with proper design, control, and ROEs robotic warriors could be more ethical and cause less collateral damage than humans.